\documentclass[letterpaper, 10 pt, conference]{ieeeconf}  

\IEEEoverridecommandlockouts                              

\title{\LARGE \bf
MCoT-RE: Multi-Faceted Chain-of-Thought and Re-Ranking for Training-Free Zero-Shot Composed Image Retrieval
}

\author{Jeong-Woo Park$^{1}$, and Seong-Whan Lee$^{1}$
\thanks{*This  research was supported by the Institute of Information \& Communications Technology Planning \& Evaluation (IITP) grant, funded by the Korea government (MSIT) (No. RS-2019-II190079, Artificial Intelligence Graduate School Program (Korea University)), and No. IITP-2025 RS-2024-00436857 (Information Technology Research Center (ITRC)).}
\thanks{$^{1}$J.-W. Park, and S.-W. Lee are with the Department of Artificial Intelligence, Korea University, Anam-dong, Seongbuk-ku, Seoul 02841, Korea.
        {\tt\small \{jeongwoo\_park, 
        sw.lee\}@korea.ac.kr}}%
}
\usepackage{graphicx}
\usepackage{booktabs}
\usepackage{multirow}
\usepackage{tabularx}
\usepackage{amsmath}
\usepackage{adjustbox}
\usepackage{cite}
\usepackage[table]{xcolor}

\usepackage{amssymb} 
\usepackage{algorithm}
\usepackage{algorithmic}

\begin{document}

\maketitle
\thispagestyle{empty}
\pagestyle{empty}


\begin{abstract}

Composed Image Retrieval (CIR) is the task of retrieving a target image from a gallery using a composed query consisting of a reference image and a modification text. Among various CIR approaches, training-free zero-shot methods based on pre-trained models are cost-effective but still face notable limitations. For example, sequential VLM-LLM pipelines process each modality independently, which often results in information loss and limits cross-modal interaction. In contrast, methods based on multimodal large language models (MLLMs) often focus exclusively on applying changes indicated by the text, without fully utilizing the contextual visual information from the reference image.
To address these issues, we propose multi-faceted Chain-of-Thought with re-ranking (MCoT-RE), a training-free zero-shot CIR framework. MCoT-RE utilizes multi-faceted Chain-of-Thought to guide the MLLM to balance explicit modifications and contextual visual cues, generating two distinct captions: one focused on modification and the other integrating comprehensive visual-textual context. The first caption is used to filter candidate images. Subsequently, we combine these two captions and the reference image to perform multi-grained re-ranking. This two-stage approach facilitates precise retrieval by aligning with the textual modification instructions while preserving the visual context of the reference image. Through extensive experiments, MCoT-RE achieves state-of-the-art results among training-free methods, yielding improvements of up to 6.24\% in Recall@10 on FashionIQ and 8.58\% in Recall@1 on CIRR.
\end{abstract}


\section{INTRODUCTION}
\label{sec:introduction}

Composed Image Retrieval (CIR) \cite{vo2019composing} enables users to retrieve a target image by combining a reference image with textual modification instructions. This paradigm provides fine-grained control over retrieval. It is particularly valuable in real-world scenarios that require understanding complex user intent, such as e-commerce~\cite{zhang2020empowering}. Accordingly, the goal of CIR is to perform precise image retrieval by accurately interpreting the user’s multimodal input. Traditionally, achieving this goal has required supervised learning on large annotated datasets~\cite{wen2021comprehensive}. While these methods achieve strong performance, they incur significant data collection and training costs. As a result, zero-shot methods that avoid task-specific training have gained attention~\cite{saito2023pic2word}. However, some zero-shot techniques, such as pseudo-token strategies \cite{ baldrati2023zero} using CLIP \cite{radford2021learning}, require auxiliary data or computational resources, motivating further exploration of training-free paradigms.

\begin{figure}[t]
    \centering
    \includegraphics[width=0.5\textwidth]{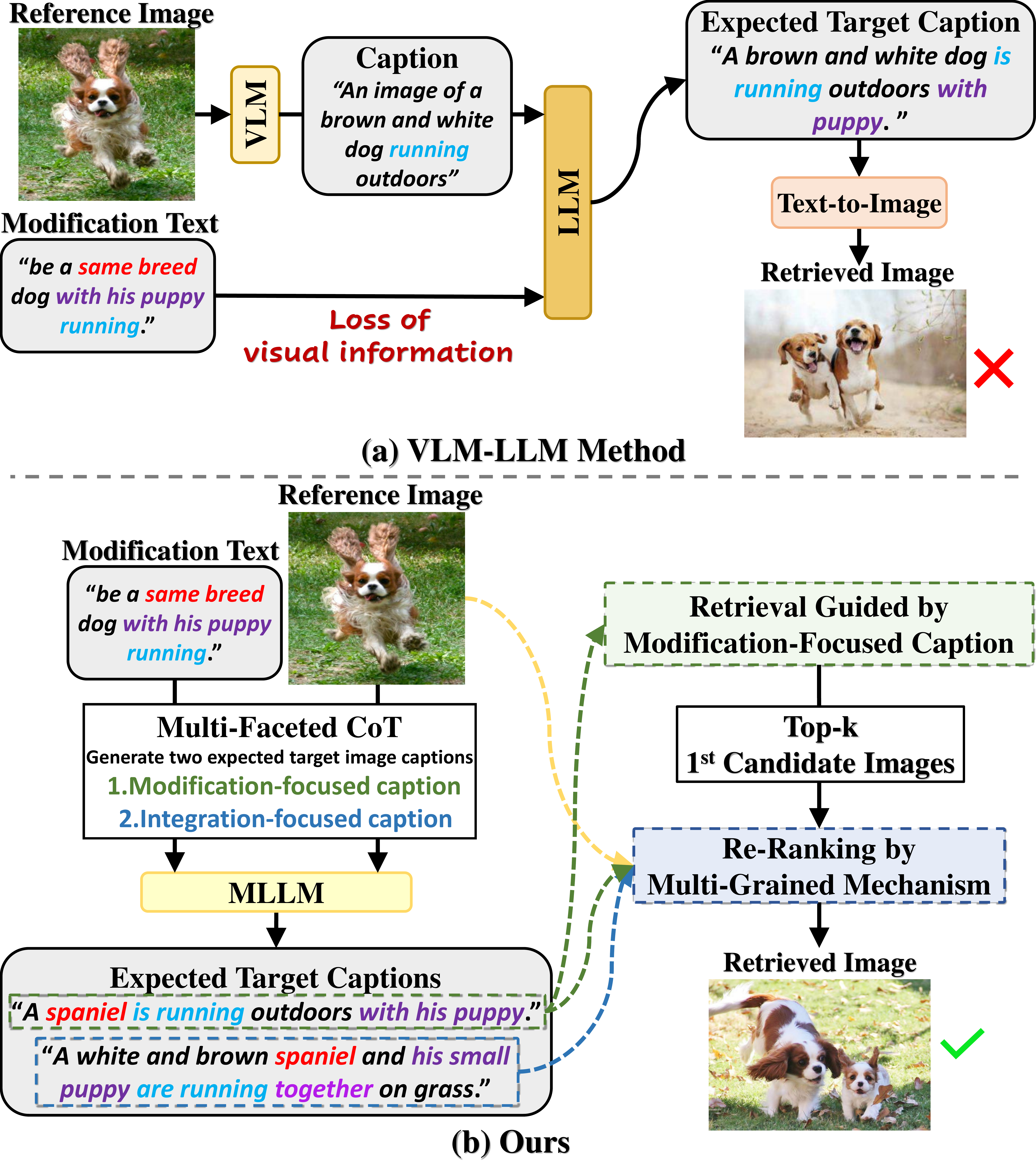}
    \caption{Comparison of existing training-free ZS-CIR method with our proposed MCoT-RE. (a) Existing methods involve visual information loss originating from sequential processing. (b) Our method performs multi-faceted reasoning, preserving both textual and visual information.}
    \label{fig:compare}
\end{figure}

A common approach to these training-free paradigms involves integrating vision-language models (VLMs) with large language models (LLMs)~\cite{karthik2023vision, wu2024training, li2024imagine}. In this framework, the VLM first generates a caption describing the reference image. This caption, combined with the modification text, is then used by the LLM to produce a target caption for retrieval. However, the VLM operates without access to the modification instructions, causing it to often omit visual details essential to the intended modification. As a result, the LLM receives incomplete information, limiting its ability to reason accurately about the desired changes. As illustrated in Fig.~\ref{fig:compare}(a), this sequential pipeline introduces early information loss that ultimately degrades retrieval performance. 

To address this limitation, multimodal large language models (MLLMs) have been introduced in CIR~\cite{karthik2023vision, bao2025mllm, yang2024ldre}. By processing image and text inputs simultaneously, MLLMs help mitigate the early information loss introduced by the sequential VLM-LLM pipeline. However, existing training-free methods based on single-pass MLLMs tend to focus narrowly on the explicitly stated modification~\cite{sun2025cotmr, tang2024reason}. These methods are typically prompted without structured reasoning strategies, limiting their ability to capture nuanced visual-textual relationships. In practice, users expect that visual aspects not mentioned in the instruction, such as background, secondary objects, style, or composition, will be preserved or adapted in a contextually appropriate manner. A focus limited to explicit instructions is misaligned with these expectations, leading to target descriptions that omit subtle but important cues and ultimately degrade retrieval accuracy.

To address these issues, we propose MCoT-RE, a training-free zero-shot CIR (ZS-CIR) framework that integrates a multi-faceted Chain-of-Thought (MCoT) process with a multi-grained re-ranking (RE) strategy. Specifically, MCoT guides the MLLM to perform multi-faceted reasoning over both the reference image and the modification text. This reasoning enables the MLLM to process explicit instructions and implicit visual context separately, ensuring that both types of cues are clearly represented and retained throughout inference. By doing so, MCoT fully exploits the MLLM’s reasoning ability while overcoming the limitations of prior methods that underutilize multimodal cues and often suffer from contextual information loss during inference.

As a result of this reasoning process, MCoT-RE produces two distinct caption types tailored for retrieval stages. The first is the modification-focused caption, describing the anticipated target image based on explicitly stated attributes in the modification instructions. The second is the integration-focused caption, which incorporates not only the explicit modifications but also contextual visual elements from the reference image. MCoT-RE systematically utilizes these captions in a two-stage retrieval pipeline. In the initial filtering stage, MCoT-RE uses the modification-focused caption to retrieve the top-\(k\) candidate images based on its similarity to each image. Next, in the multi-grained re-ranking stage, MCoT-RE precisely adjusts the ranking by comparing these candidates against a combination of both captions and the reference image. By jointly leveraging explicit visual signals corresponding to the user’s intended changes and implicit contextual cues often underutilized in previous methods, MCoT-RE enhances the precision of the final retrieval results. An overview of this reasoning-guided generation and retrieval pipeline is shown in Fig.~\ref{fig:compare}(b).

In conclusion, MCoT-RE substantially improves ZS-CIR performance, achieving state-of-the-art results on standard benchmarks such as FashionIQ~\cite{wu2021fashion} and CIRR~\cite{liu2021image}. Specifically, it outperforms existing training-free ZS-CIR methods, achieving higher Recall@10 on FashionIQ and higher Recall@1 on CIRR.

The key contributions of this study are as follows:
\begin{itemize}
\item We propose MCoT-RE, a training-free framework for ZS-CIR that incorporates a multi-faceted CoT module to leverage MLLM capabilities and a multi-grained re-ranking mechanism for accurate composed image retrieval.

\item By using an MLLM guided by our MCoT, we address limitations in leveraging the input modalities and model inference capabilities. This approach generates distinct modification-focused captions for initial filtering and integration-focused captions for a subsequent re-ranking stage.

\item MCoT-RE outperforms previous state-of-the-art training-free methods on the FashionIQ and CIRR by up to 6.24\% on Recall@10 and 8.58\% on Recall@1.
\end{itemize}

\section{RELATED WORKS}

\subsection{Zero‑Shot Composed Image Retrieval}

ZS-CIR retrieves a target image using a reference image and a textual modification without task-specific training. Training-free methods are typically categorized as either VLM-LLM pipelines or direct MLLM approaches. In VLM-LLM pipelines, a VLM generates a caption from the reference image, which is then revised by an LLM using the modification text \cite{yang2024ldre, li2024imagine, karthik2023vision}. Because the VLM operates without knowledge of the modification, it often misses important visual details. This sequential setup also introduces latency and can propagate early errors. In contrast, direct MLLMs process the image and text jointly to produce a single caption \cite{bao2025mllm}. Although this reduces modality separation, these models tend to focus on the explicit instruction while neglecting implicit visual context such as background, style, or layout. To address these limitations, our MCoT-RE guides the MLLM to generate two distinct captions. One focuses on the modification described in the instruction, while the other incorporates contextual information based on both the instruction and the reference image. These outputs are used in a two-stage retrieval process that first filters candidates based on the explicitly requested change and then refines them using both types of information.

\begin{figure*}[htbp]
    \centering
    \includegraphics[width=\textwidth]{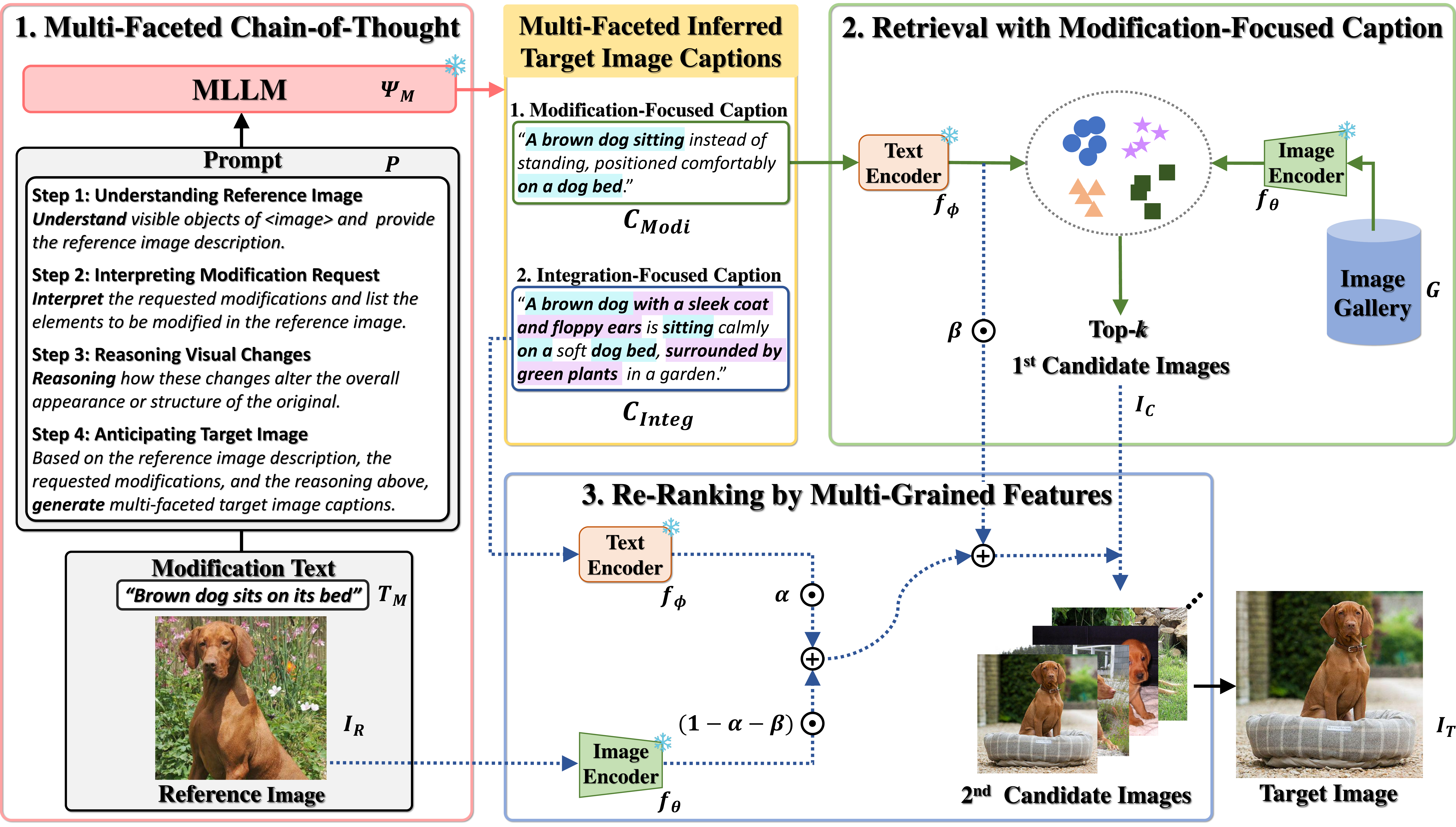}
    \caption{Overall pipeline of MCoT-RE. Our framework employs an MCoT process using an MLLM to jointly reason over the reference image and modification text. It generates two captions: a modification-focused caption for candidate filtering and an integration-focused caption for multi-grained re-ranking. By combining these captions with the reference image, MCoT-RE enables multi-grained retrieval in a training-free manner.}
    \label{fig:overview}
\end{figure*}

\subsection{Chain-of-Thought Prompting in Composed Image Retrieval}

CoT prompting enhances an LLM’s reasoning by generating explicit, step-by-step inferences. Initially successful in tasks like math and commonsense reasoning, CoT has recently been applied to multimodal LLMs \cite{sun2025cotmr, tang2024reason}, which combine visual and textual inputs for more expressive reasoning in CIR. Most existing CoT-based CIR methods generate a single caption that focuses primarily on the modification text, with little regard for contextual cues from the reference image. This single caption is used directly for retrieval without incorporating visual information beyond the explicitly stated elements in the instruction. As a consequence, implicit visual cues from the reference image, such as style, background, and spatial composition, are often ignored. To address these limitations, our MCoT-RE adopts an MCoT in which the model generates two distinct captions. One caption reflects the explicit modification described in the instruction, while the other incorporates contextual information based on both the instruction and the reference image. These captions convey complementary information, enabling the model to utilize both directive and contextual cues when performing image retrieval.


\section{METHOD}
\label{sec:method} 

\subsection{Problem Formulation}
\label{subsec:problem_formulation}

Given a reference image \(I_R\) and a text-based modification instruction \(T_M\), CIR ranks a gallery of \(N\) images \(G = \{I_j\}_{j=1}^{N}\).
The objective is to place the target image \(I_T\), which reflects the visual changes described in \(T_M\) while preserving the essential visual attributes of \(I_R\), at the top of the list. We address this task in a training-free and zero-shot setting by leveraging an MLLM. Our core approach utilizes an MLLM, denoted as \(\Psi_{M}\), guided by our proposed MCoT prompting strategy.
The MCoT process generates two descriptions based on the composed query: a modification-focused caption \(C_{\mathit{Modi}}\) and an integration-focused caption \(C_{\mathit{Integ}}\).
These captions are then employed in a two-stage retrieval and re-ranking framework to rank candidate images such that those aligned with the intended modification appear at the top, as illustrated in Fig.~\ref{fig:overview}.

\subsection{Multi-Faceted Chain-of-Thought}
\label{subsec:mcot}

To fully leverage the reasoning capabilities of an MLLM for CIR, we propose the MCoT prompting method. MCoT encourages the MLLM to analyze the reference image and modification text jointly from both explicit and implicit perspectives. This approach ensures a comprehensive consideration of both explicit modification instructions and the implicit visual context. The MCoT process can be formally represented as:
\begin{equation} \label{eq:mcot_process}
\{C_{\mathit{Modi}}, C_{\mathit{Integ}}\} = \Psi_M(I_R, T_M, P),
\end{equation}
where \(P\) is the prompt specifically designed to elicit step-by-step reasoning, yielding both a modification-focused caption \(C_{\mathit{Modi}}\) and an integration-focused caption \(C_{\mathit{Integ}}\).

\subsubsection{Step-by-step Reasoning Process}
\label{ssubsec:mcot_steps}

MCoT guides \(\Psi_M\) through a structured, four-step reasoning process embedded within the prompt \(P\), fostering a deep understanding of the reference image and modification text.

\textbf{Step 1. Understanding Reference Image:} \(\Psi_M\) begins by meticulously analyzing the key visual elements of the reference image \(I_R\), including primary objects, their attributes, background details, overall composition, and stylistic features. The goal is to establish a comprehensive understanding of the image's intrinsic properties, independent of \(T_M\).

\textbf{Step 2. Interpreting Modification Request:} Next, \(\Psi_M\) carefully parses the \(T_M\) to identify which visual elements are targeted for change and the nature of the requested modification. This involves pinpointing specific entities or attributes and the direction of the desired transformation.

\textbf{Step 3. Reasoning Visual Changes:} In the third step, \(\Psi_M\) reasons about the implications of the identified modifications on the overall visual appearance of the scene depicted in \(I_R\). This step is crucial for inferring potential unintended changes and, importantly, identifying elements that should remain invariant despite the modifications.

\begin{figure}[!t]
    \centering
\includegraphics[width=0.485\textwidth]{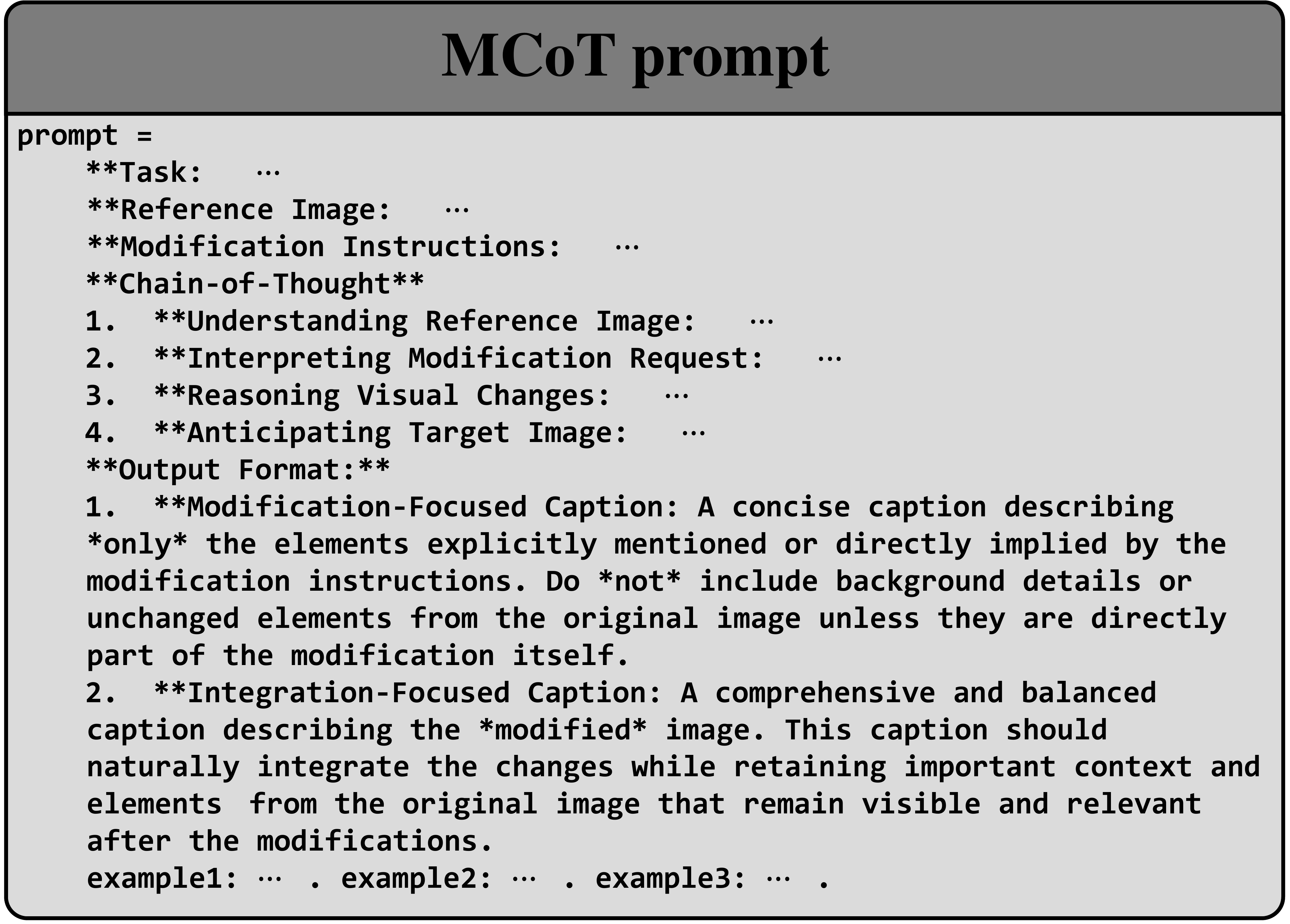}
    \caption{Prompt template for generating the two target captions. The figure highlights the instructions guiding the MLLM to produce $C_{\mathit{Modi}}$ and $C_{\mathit{Integ}}$, with other sections abbreviated.}
    \label{fig:MCoT_prompt}
\end{figure}

\textbf{Step 4. Anticipating Target Image:} Finally, integrating the insights from the preceding steps, $\Psi_M$ generates multi-faceted descriptions of the target image. A key aspect of this stage is the generation of two captions, $C_{\mathit{Modi}}$ and $C_{\mathit{Integ}}$, reflecting different facets of the reasoning process, as illustrated in the prompt template shown in Fig.~\ref{fig:MCoT_prompt}.

This explicit step-by-step structure within the prompt \(P\) ensures that \(\Psi_M\) develops a nuanced understanding before generating the target captions.

\subsubsection{Multi-Faceted Target Captions}
\label{ssubsec:mcot_captions}

A core innovation of MCoT lies in generating two distinct target captions from a single MLLM inference. 
The first caption, \(C_{\mathit{Modi}}\), explicitly describes only the changes specified in the modification text \(T_M\), and is used as a query to retrieve initial candidate images matching these changes. 
The second caption, \(C_{\mathit{Integ}}\), extends this by incorporating preserved visual context from the reference image \(I_R\), including elements such as background, style, and ambiance. This integration-focused caption, along with the modification-focused caption and the reference image features, is subsequently utilized in a re-ranking stage to refine the initial candidate selection by capturing both explicit modifications and implicit visual context.

\subsection{Retrieval Guided by Modification-Focused Target Caption}
\label{subsec:retrieval_stage1}

The modification-focused caption \(C_{\mathit{Modi}}\), generated using MCoT, is employed in the initial candidate retrieval stage. 
In this stage, \(C_{\mathit{Modi}}\) is converted into a text embedding vector \(F_{\mathit{Modi}}\) using a pre-trained text encoder \(f_\phi\). Concurrently, all images \(I_j\) in the gallery \(G\) are encoded into image embeddings using a corresponding pre-trained image encoder \(f_\theta\). The cosine similarity between the caption embedding and each image embedding is then computed as follows:
\begin{equation} 
\label{eq:stage1_sim}
S_{1^{st}}(I_j) = \mathit{sim}(F_{\mathit{Modi}}, f_\theta(I_j)), \quad \forall I_j \in G.
\end{equation}
Here, \(sim\) denotes the cosine similarity, computed by first \(\ell_2\)-normalizing each input vector and then taking their dot product to measure angular closeness. Based on these similarity scores \(S_{1^{st}}\), the top-\(k\) candidate images \(I_C = \{I_c^1, I_c^2, \ldots, I_c^k\}\) are selected as:
\begin{equation} \label{eq:stage1_selection}
I_C = \bigl\{\,I_j \in G \mid S_{1^{st}}(I_j)\ \text{is among top-}k\ \text{in}\ G \bigr\}.
\end{equation}

This step narrows down the gallery to a smaller set of images closely aligned with the explicit modification request.

\begin{algorithm}[t]
\caption{Retrieval process of MCoT-RE}
\label{alg:mcot-re}
\begin{algorithmic}[1]

\STATE \textbf{Input:} reference image $I_R$, modification text $T_M$, MCoT prompt $P$, image gallery $G$
\STATE \textbf{Parameters:} frozen MLLM $\Psi_M$, text encoder $f_\phi$, image encoder $f_\theta$, number of top candidates $k$, weights $\alpha, \beta$
\STATE \textbf{Output:} ranked candidate list $I_C^*$

\STATE Initialize $\Psi_M, f_\phi, f_\theta$
\STATE Generate two captions:
\STATE \hspace{\algorithmicindent} $\{C_{\mathit{Modi}},\,C_{\mathit{Integ}}\} \leftarrow \Psi_M(I_R,\,T_M,\,P)$
\STATE Compute embeddings:
\STATE \hspace{\algorithmicindent} $F_{\mathit{Modi}} \leftarrow f_\phi(C_{\mathit{Modi}})$
\STATE \hspace{\algorithmicindent} $F_{\mathit{Integ}} \leftarrow f_\phi(C_{\mathit{Integ}})$
\STATE \hspace{\algorithmicindent} $F_R \leftarrow f_\theta(I_R)$

\FOR{each image $I_j$ in $G$}
    \STATE $S_{1^{st}}(I_j) \leftarrow sim\bigl(F_{\mathit{Modi}},\,f_\theta(I_j)\bigr)$
\ENDFOR
\STATE $I_C \leftarrow \bigl\{\,I_j \in G \mid S_{1^{st}}(I_j)\ \text{is among the } k \text{ largest in } G \bigr\}$

\STATE Compute combined feature:
\STATE \hspace{\algorithmicindent} $F_{\mathit{comb}} \leftarrow \alpha\,F_{\mathit{Modi}} + \beta\,F_{\mathit{Integ}} + (1-\alpha-\beta)\,F_R$

\FOR{each candidate $I_c^m$ in $I_C$}
    \STATE $S_{2^{nd}}(I_c^m) \leftarrow sim\bigl(F_{\mathit{comb}},\,f_\theta(I_c^m)\bigr)$
\ENDFOR
\STATE $I_C^* \leftarrow \operatornamewithlimits{argsort}(-S_{2^{nd}}(I_c^m))$
\RETURN $I_C^*$

\end{algorithmic}
\end{algorithm}

\subsection{Re-Ranking by Multi-Grained Features}
\label{subsec:retrieval_stage2}

Although the first stage selects the top-\(k\) candidates \(I_C\) based on the explicit modifications captured by \(C_{\mathit{Modi}}\), it often overlooks the implicit visual context of \(I_R\).
To mitigate this limitation, we introduce a context-guided re-ranking step that jointly leverages \(C_{\mathit{Integ}}\), \(C_{\mathit{Modi}}\), and \(I_R\). Here, the integration-focused caption embedding is defined as \(F_{\mathit{Integ}} = f_\phi(C_{\mathit{Integ}})\), and the reference-image embedding as \(F_R = f_\theta(I_R)\). A combined query feature \(F_{\mathit{comb}}\) is constructed by a weighted fusion of the corresponding embeddings:
\begin{equation} \label{eq:stage2_combine}
F_{\mathit{comb}}
  = \alpha\,F_{\mathit{Modi}}
  + \beta\,F_{\mathit{Integ}}
  + (1-\alpha-\beta)\,F_R,
\end{equation}
where \(\alpha, \beta \in [0, 1]\) and \(\alpha + \beta \le 1\) are balancing hyperparameters.
The cosine similarity score between \(F_{\mathit{comb}}\) and each candidate image is computed as:
\begin{equation} \label{eq:stage2_sim}
S_{2^{nd}}\!\bigl(I_c^m\bigr)
  = \mathit{sim} \bigl(F_{\mathit{comb}},\,f_\theta(I_c^m)\bigr),
  \quad \forall\,I_c^m \in I_C.
\end{equation}

The candidates are then sorted in the descending order of \(S_{2^{nd}}\) to yield the final ranking:
\begin{equation} \label{eq:final_selection}
I_C^{*}
  = \operatornamewithlimits{argsort}_{I_c^m \in I_C}
    \bigl(-\,S_{2^{nd}}(I_c^m)\bigr),
\end{equation}
where $\operatornamewithlimits{argsort}$ returns the indices that sort inputs in ascending order, and the minus sign reverses this order so that candidates with higher $S_{2^{nd}}$ values are ranked earlier.
This multi-grained strategy enables nuanced comparison among candidates and preserves the essential visual attributes of \(I_R\).

The complete MCoT-RE pipeline, from caption generation to final retrieval, is summarized in Algorithm~\ref{alg:mcot-re}.

\begin{table}[!t]
\centering
\caption{R@K results on FashionIQ. The best scores for each backbone are highlighted in \textbf{bold}.}
\label{tab:fashioniq_r_at_k}
\adjustbox{width=0.49\textwidth}{%
{%
\fontsize{9pt}{11.6pt}\selectfont
\begin{tabular}{lcccccc}
\toprule
\multicolumn{1}{c}{\textbf{Backbone /}} & \multicolumn{2}{c}{\textbf{Shirt}} & \multicolumn{2}{c}{\textbf{Dress}} & \multicolumn{2}{c}{\textbf{Toptee}} \\
\cmidrule(lr){2-7}
\multicolumn{1}{c}{\textbf{Method}} & \textbf{R@10} & \textbf{R@50} & \textbf{R@10} & \textbf{R@50} & \textbf{R@10} & \textbf{R@50} \\
\midrule
\multicolumn{7}{c}{\textit{\textbf{Zero-shot methods that need an additional pre-training step}}} \\
\midrule
\multicolumn{1}{c}{ViT-L-14 /} & & & & & & \\
CompoDiff \cite{gu2024compodiff}  & \textbf{37.69} & \textbf{49.08} & \textbf{32.24} & \textbf{46.27} & \textbf{38.12} & \textbf{50.57} \\
LinCIR \cite{gu2024language}     & 29.10        & 46.81 & 20.92        & 42.44 & 28.81        & 50.18 \\
\midrule
\multicolumn{1}{c}{ViT-G-14 /} & & & & & & \\
Pic2Word \cite{saito2023pic2word}  & 33.17        & 50.39 & 25.43        & 47.65 & 35.24        & 57.62 \\
CompoDiff \cite{gu2024compodiff}  & 41.31        & 55.17 & 37.78        & 49.10 & 44.26        & 56.41 \\
LinCIR \cite{gu2024language}  & \textbf{46.76} & \textbf{65.11} & \textbf{38.08} & \textbf{60.88} & \textbf{50.48} & \textbf{71.09} \\
\midrule
\multicolumn{7}{c}{\textit{\textbf{Training-free zero-shot methods}}} \\
\midrule
\multicolumn{1}{c}{ViT-L-14 /} & & & & & & \\
WeiMoCIR \cite{wu2024training}       & 32.78        & 48.97 & 25.88        & 47.30 & 35.95 & 56.71 \\
LDRE \cite{yang2024ldre}      & 31.04        & 51.22 & 22.93  & 46.76 & 31.57        & 53.64 \\
OSrCIR \cite{tang2024reason}       & 33.17        &  52.03 & 29.70        &  51.81 &  36.92 & 59.27 \\ 
\rowcolor{gray!20} \textbf{Ours}  & \textbf{36.60} & \textbf{53.09} & \textbf{35.94} & \textbf{55.82} & \textbf{43.55}        & \textbf{64.30} \\
\midrule
\multicolumn{1}{c}{ViT-G-14 /} & & & & & & \\
WeiMoCIR \cite{wu2024training}      & 37.73 & 56.18 & 30.99 & 52.45 & 42.38 & 63.23 \\
LDRE \cite{yang2024ldre}      & 35.94        & 58.58 & 26.11  & 51.12 & 35.42        & 56.67 \\
OSrCIR \cite{tang2024reason}   & 38.65        &  54.71 & 33.02     &  54.78 &  41.04 & 61.83 \\ 
\rowcolor{gray!20} \textbf{Ours}  & \textbf{42.35} & \textbf{59.81} & \textbf{34.51} & \textbf{56.67} & \textbf{45.74} & \textbf{67.57} \\
\bottomrule
\end{tabular}
}%
}
\end{table}


\section{EXPERIMENTS}\label{sec:experiment}

\subsection{Datasets and Evaluation Protocol.}
We evaluate MCoT-RE on FashionIQ and CIRR. FashionIQ contains three categories (Shirt, Dress, and Toptee), with 46k training and 15k validation/test images. We report results on the validation split, as the method is training-free. CIRR comprises more than 36k images with complex textual modifications, and we evaluate it on the official test split only. Performance is measured using Recall@K (R@K). For CIRR we additionally report Recall$_{\textit{subset}}$@K ({R\(_{s}\)}@K) on visually similar subsets.

\subsection{Implementation Details.}

Our framework operates entirely in a training-free manner, obviating the need for task‑specific fine‑tuning~\cite{lee2020uncertainty, lee2001automatic}. We employ pre‑trained encoders (CLIP‑based backbones) to obtain visual and textual embeddings~\cite{ lee1995multilayer, fujisawa1999information}. For each dataset, we generate the anticipated target caption using a structured prompt‑chaining procedure with the Gemini~\cite{team2023gemini} 1.5. The fusion weights in Eq.~(\ref{eq:stage2_combine}) are fixed to $\alpha=0.05$ and $\beta=0.9$ for all experiments. We empirically determine the optimal number of initial candidates \(k\) for re-ranking to be 150 for FashionIQ and 200 for CIRR, balancing computational efficiency with retrieval accuracy.

\begin{table}[!t]
\centering
\caption{R@K results on CIRR. The best scores for each backbone are highlighted in \textbf{bold}.}
\label{tab:CIRR_results}
\adjustbox{width=0.49\textwidth}{%
{%
\fontsize{9pt}{11pt}\selectfont
\begin{tabular}{lcccccc}
\toprule
\multicolumn{1}{c}{\textbf{Backbone /}} & \multicolumn{3}{c}{\textbf{Recall@K}} & \multicolumn{3}{c}{\textbf{Recall$_{\textit{subset}}$@K}} \\
\cmidrule(lr){2-7}
\multicolumn{1}{c}{\textbf{Method}} & \textbf{R@1} & \textbf{R@5} & \textbf{R@10} & \textbf{R$_{\textit{s}}$@1} & \textbf{R$_{\textit{s}}$@2} & \textbf{R$_{\textit{s}}$@3} \\
\midrule
\multicolumn{7}{c}{\textit{\textbf{Zero-shot methods that need an additional pre-training step}}} \\
\midrule
\multicolumn{1}{c}{ViT-L-14 /} & & & & & & \\
Pic2Word \cite{saito2023pic2word}  & 23.90        & 51.70 & 65.30 & 53.76 & 74.46 & 87.08 \\
CompoDiff \cite{gu2024compodiff} & 18.24        & 53.14 & \textbf{70.82} & \textbf{57.42} & 77.10 & 87.90 \\
LinCIR \cite{gu2024language}    & \textbf{25.04}        & \textbf{53.25} & 66.68     & 57.11 & \textbf{77.37} & \textbf{88.89} \\
\midrule
\multicolumn{1}{c}{ViT-G-14 /} & & & & & & \\
Pic2Word \cite{saito2023pic2word}  & 30.41        & 58.12 & 69.23 & \textbf{68.92} & \textbf{85.45} & 93.04 \\
SEARLE \cite{baldrati2023zero}    & 34.80        & 64.07 & 75.11 & 68.72 & 84.70 & \textbf{93.23} \\
CompoDiff \cite{gu2024compodiff} & 26.71        & 55.14 & 74.52 & 64.54 & 82.39 & 91.81 \\
LinCIR \cite{gu2024language}    & \textbf{35.25} & \textbf{64.72} & \textbf{76.05} & 63.35 & 82.22 & 91.98 \\
\midrule
\multicolumn{7}{c}{\textit{\textbf{Training-free zero-shot methods}}} \\
\midrule
\multicolumn{1}{c}{ViT-L-14 /} & & & & & & \\
WeiMoCIR \cite{wu2024training}  & 30.94        & 60.87 & 73.08 & 58.55 & 79.06 & 90.07 \\
LDRE \cite{yang2024ldre}      & 26.53        & 55.57 & 67.54 & 60.43 & 80.31 & 89.90 \\
OSrCIR \cite{tang2024reason} & 29.45        & 57.68 & 69.86 & 62.12 & 81.92 & 91.10 \\
\rowcolor{gray!20} \textbf{Ours}       & \textbf{39.52} & \textbf{69.18} & \textbf{79.28} & \textbf{70.19} & \textbf{86.34} & \textbf{93.83} \\
\midrule
\multicolumn{1}{c}{ViT-G-14 /} & & & & & & \\
WeiMoCIR \cite{wu2024training}  & 31.04        & 60.41 & 72.27 & 58.84 & 78.92 & 89.64 \\
LDRE \cite{yang2024ldre}      & 36.15        & 66.39 & 77.25 & 68.82 & 85.66 & 93.76 \\
OSrCIR \cite{tang2024reason} & 37.26        & 67.25 & 77.33 & 69.22 & 85.28 & 93.55 \\
\rowcolor{gray!20} \textbf{Ours} & \textbf{39.37} & \textbf{68.92} & \textbf{79.30} & \textbf{70.82} & \textbf{86.92} & \textbf{93.88} \\
\bottomrule
\end{tabular}
}%
}
\end{table}

\subsection{Comparison with State-of-the-Art Methods}

As shown in Tab.~\ref{tab:fashioniq_r_at_k} and Tab.~\ref{tab:CIRR_results}, MCoT-RE outperforms existing training-free baselines across both FashionIQ and CIRR. It achieves up to 6.63\% higher Recall@10 on FashionIQ and up to 8.58\% higher Recall@1 on CIRR compared to the strongest baselines. CIRR presents retrieval challenges through the inclusion of visually similar non-target candidates and often underspecified modification instructions, which complicate the task of retrieving the correct image in realistic settings. Despite these conditions, MCoT-RE surpasses all baselines across different CLIP backbones, achieving state-of-the-art results, particularly in Recall@1, as well as across other standard metrics and subset-based evaluations. These results demonstrate the robustness of MCoT-RE, which jointly utilizes both directive and contextual information to address the challenges of CIRR.

\subsection{Qualitative Results}
\begin{figure}[!t]
    \centering
\includegraphics[width=0.485\textwidth]{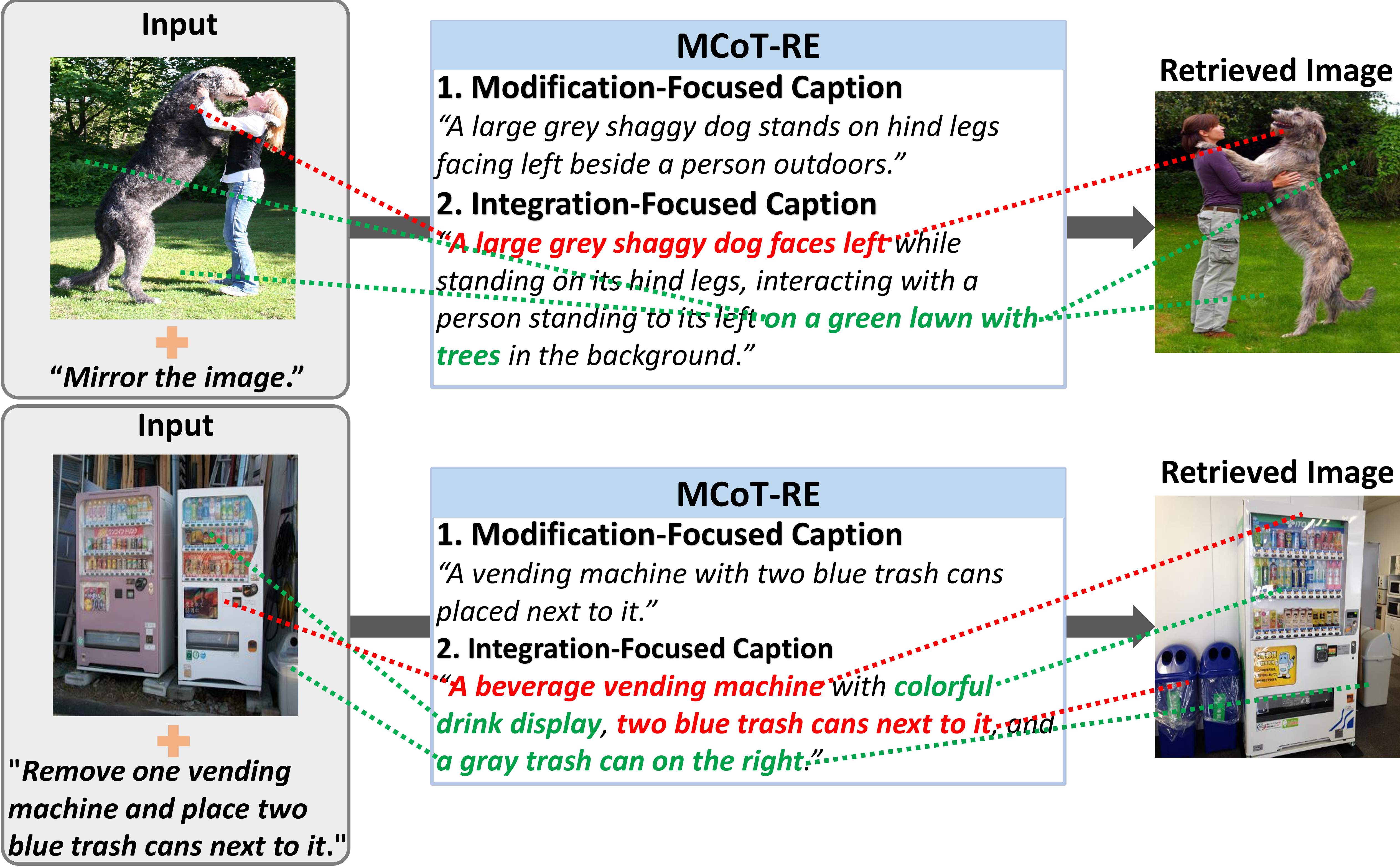}
    \caption{Successful retrieval examples with MCoT-RE from CIRR.}
    \label{fig:real}
\end{figure}

Fig.~\ref{fig:real} shows successful retrieval examples with MCoT-RE. Given the instruction ``Mirror the image", our method accurately retrieves a mirrored image preserving context. For the instruction ``Remove one vending machine and place two blue trash cans next to it", the model correctly captures explicit changes and implicit context. These examples demonstrate effective multi-faceted reasoning.

\subsection{Ablation Study}

To assess the effectiveness of each component in our framework, we conduct ablation studies on both FashionIQ and CIRR. As shown in Tab.~\ref{tab:ablation_study}, we evaluate the model without the initial filtering step (w/o 1st Filtering), without the re-ranking step (w/o Re-Ranking), and with only the modification-focused or integration-focused captions. Removing the initial filtering step or re-ranking leads to performance decreases of up to 9.58\% on FashionIQ and 5.44\% on CIRR. The largest performance drop is observed when using only the integration-focused caption, reaching up to 12.90\% on FashionIQ and 17.65\% on CIRR. These results demonstrate that both filtering and re-ranking contribute to performance, and that the combination of captions is more effective than using either in isolation.

\begin{table}[!t]
\centering
\caption{Ablation study results on FashionIQ (shirt, dress, and toptee) and CIRR. All experiments are performed with the ViT-G-14 CLIP model.}
\label{tab:ablation_study}
\adjustbox{width=0.475\textwidth}{%
{%
\begin{tabular}{lccc|cc}
\toprule
\multirow{2}{*}{\textbf{Method}} & \textbf{Shirt} & \textbf{Dress} & \textbf{Toptee} & \multicolumn{2}{c}{\textbf{CIRR}} \\
\cmidrule(lr){2-4} \cmidrule(lr){5-6} 
& \textbf{R@10} & \textbf{R@10} & \textbf{R@10} & \textbf{R@5} & \textbf{R$_{\textit{s}}$@1} \\
\midrule
w/o 1st Filtering      & 34.73 & 32.77 & 39.97 & 65.61 & 66.83 \\
w/o Re-Ranking         & 35.44 & 30.47 & 41.23 & 65.38 & 66.37 \\
\textit{C\textsubscript{Modi}} only  & 37.80 & 32.11 & 39.46 & 67.21 & 64.82 \\
\textit{C\textsubscript{Integ}} only & 29.45 & 27.12 & 34.23 & 61.03 & 53.17 \\
\rowcolor{gray!20} \textbf{Ours}          & \textbf{42.35} & \textbf{34.51} & \textbf{45.74} & \textbf{68.92} & \textbf{70.82} \\
\bottomrule
\end{tabular}%
}
}
\end{table}

\section{CONCLUSION}

In this work, we propose MCoT-RE, a training-free ZS-CIR framework addressing existing limitations in comprehensively leveraging both explicit modifications and implicit visual context. By employing an MCoT process, MCoT-RE guides the MLLM in reasoning to generate distinct captions tailored for a two-stage retrieval involving initial filtering and subsequent multi-grained re-ranking. This method effectively integrates explicit textual modifications with implicit visual information. Comprehensive evaluations on the FashionIQ and CIRR demonstrate that MCoT-RE outperforms state-of-the-art methods, confirming its effectiveness for accurate composed image retrieval. For future work, we plan to explore hierarchical attribute reasoning that decomposes visual elements into low-level (color, texture), mid-level (pattern, shape), and high-level (style, composition) features with dedicated reasoning processes for each layer~\cite{lee1996multiresolution}.

\bibliographystyle{IEEEtran}
\bibliography{references}

\begin{thebibliography}{10}
\providecommand{\url}[1]{#1}
\csname url@samestyle\endcsname
\providecommand{\newblock}{\relax}
\providecommand{\bibinfo}[2]{#2}
\providecommand{\BIBentrySTDinterwordspacing}{\spaceskip=0pt\relax}
\providecommand{\BIBentryALTinterwordstretchfactor}{4}
\providecommand{\BIBentryALTinterwordspacing}{\spaceskip=\fontdimen2\font plus
\BIBentryALTinterwordstretchfactor\fontdimen3\font minus \fontdimen4\font\relax}
\providecommand{\BIBforeignlanguage}[2]{{%
\expandafter\ifx\csname l@#1\endcsname\relax
\typeout{** WARNING: IEEEtran.bst: No hyphenation pattern has been}%
\typeout{** loaded for the language `#1'. Using the pattern for}%
\typeout{** the default language instead.}%
\else
\language=\csname l@#1\endcsname
\fi
#2}}
\providecommand{\BIBdecl}{\relax}
\BIBdecl

\bibitem{vo2019composing}
N.~Vo \emph{et~al.}, ``Composing text and image for image retrieval-an empirical odyssey,'' in \emph{Proc. IEEE/CVF Conf. Comput. Vis. Pattern Recognit. (CVPR)}, 2019, pp. 6439--6448.

\bibitem{zhang2020empowering}
J.~Zhang and D.~Tao, ``Empowering things with intelligence: a survey of the progress, challenges, and opportunities in artificial intelligence of things,'' \emph{IEEE Internet Things J.}, vol.~8, pp. 7789--7817, 2020.

\bibitem{wen2021comprehensive}
H.~Wen, X.~Song, X.~Yang, Y.~Zhan, and L.~Nie, ``Comprehensive linguistic-visual composition network for image retrieval,'' in \emph{Proc. 44th Int. ACM SIGIR Conf. Res. Dev. Inf. Retr. (SIGIR)}, 2021, pp. 1369--1378.

\bibitem{saito2023pic2word}
K.~Saito \emph{et~al.}, ``{Pic2Word}: mapping pictures to words for zero-shot composed image retrieval,'' in \emph{Proc. IEEE/CVF Conf. Comput. Vis. Pattern Recognit. (CVPR)}, 2023, pp. 19\,305--19\,314.

\bibitem{baldrati2023zero}
A.~Baldrati, L.~Agnolucci, M.~Bertini, and A.~Del~Bimbo, ``Zero-shot composed image retrieval with textual inversion,'' in \emph{Proc. IEEE/CVF Int. Conf. Comput. Vis. (ICCV)}, 2023, pp. 15\,338--15\,347.

\bibitem{radford2021learning}
A.~Radford \emph{et~al.}, ``Learning transferable visual models from natural language supervision,'' in \emph{Int. Conf. Mach. Learn. (ICML)}, 2021, pp. 8748--8763.

\bibitem{karthik2023vision}
S.~Karthik, K.~Roth, M.~Mancini, and Z.~Akata, ``Vision-by-language for training-free compositional image retrieval,'' \emph{arXiv preprint arXiv:2310.09291}, 2023.

\bibitem{wu2024training}
R.-D. Wu, Y.-Y. Lin, and H.-F. Yang, ``Training-free zero-shot composed image retrieval via weighted modality fusion and similarity,'' \emph{arXiv preprint arXiv:2409.04918}, 2024.

\bibitem{li2024imagine}
Y.~Li, F.~Ma, and Y.~Yang, ``Imagine and seek: Improving composed image retrieval with an imagined proxy,'' \emph{arXiv preprint arXiv:2411.16752}, 2024.

\bibitem{bao2025mllm}
T.~Bao, C.~Liu, D.~Xu, Z.~Zheng, and T.~Xu, ``{MLLM-I2W}: Harnessing multimodal large language model for zero-shot composed image retrieval,'' in \emph{Proc. of the 31st Int. Conf. on Comput. Linguistics (COLING)}, 2025, pp. 1839--1849.

\bibitem{yang2024ldre}
Z.~Yang, D.~Xue, S.~Qian, W.~Dong, and C.~Xu, ``{LDRE}: {LLM}-based divergent reasoning and ensemble for zero-shot composed image retrieval,'' in \emph{Proc. Int. ACM SIGIR Conf. Res. Dev. Inf. Retr. (SIGIR)}, 2024, pp. 80--90.

\bibitem{sun2025cotmr}
Z.~Sun, D.~Jing, and Z.~Lu, ``{CoTMR}: chain-of-thought multi-scale reasoning for training-free zero-shot composed image retrieval,'' \emph{arXiv preprint arXiv:2502.20826}, 2025.

\bibitem{tang2024reason}
Y.~Tang \emph{et~al.}, ``Reason-before-retrieve: One-stage reflective chain-of-thoughts for training-free zero-shot composed image retrieval,'' \emph{arXiv preprint arXiv:2412.11077}, 2024.

\bibitem{wu2021fashion}
H.~Wu \emph{et~al.}, ``Fashion{IQ}: A new dataset towards retrieving images by natural language feedback,'' in \emph{Proc. IEEE/CVF Conf. Comput. Vis. Pattern Recognit. (CVPR)}, 2021, pp. 11\,307--11\,317.

\bibitem{liu2021image}
Z.~Liu, C.~Rodriguez-Opazo, D.~Teney, and S.~Gould, ``Image retrieval on real-life images with pre-trained vision-and-language models,'' in \emph{Proc. IEEE/CVF Int. Conf. Comput. Vis. (ICCV)}, 2021, pp. 2125--2134.

\bibitem{gu2024compodiff}
G.~Gu \emph{et~al.}, ``{CompoDiff}: Versatile composed image retrieval with latent diffusion,'' \emph{Trans. Mach. Learn. Res.}, 2024.

\bibitem{gu2024language}
G.~Gu, S.~Chun, W.~Kim, Y.~Kang, and S.~Yun, ``Language-only training of zero-shot composed image retrieval,'' in \emph{Proc. IEEE/CVF Conf. Comput. Vis. Pattern Recognit. (CVPR)}, 2024, pp. 13\,225--13\,234.

\bibitem{lee2020uncertainty}
G.-H. Lee and S.-W. Lee, ``Uncertainty-aware mesh decoder for high fidelity 3d face reconstruction,'' in \emph{Proc. IEEE/CVF Con. Comput. Vis. Pattern Recognit. (CVPR)}, 2020, pp. 6100--6109.

\bibitem{lee2001automatic}
M.-S. Lee, Y.-M. Yang, and S.-W. Lee, ``Automatic video parsing using shot boundary detection and camera operation analysis,'' \emph{Pattern Recognit.}, vol.~34, no.~3, pp. 711--719, 2001.

\bibitem{lee1995multilayer}
S.-W. Lee, ``Multilayer cluster neural network for totally unconstrained handwritten numeral recognition,'' \emph{Neural Networks}, vol.~8, no.~5, pp. 783--792, 1995.

\bibitem{fujisawa1999information}
H.~Fujisawa, H.~Sako, Y.~Okada, and S.-W. Lee, ``Information capturing camera and developmental issues,'' in \emph{Proc. Int. Conf. Document Anal. Recognit.}, 1999, pp. 205--208.

\bibitem{team2023gemini}
G.~Team \emph{et~al.}, ``Gemini: a family of highly capable multimodal models,'' \emph{arXiv preprint arXiv:2312.11805}, 2023.

\bibitem{lee1996multiresolution}
S.-W. Lee, C.-H. Kim, H.~Ma, and Y.~Y. Tang, ``Multiresolution recognition of unconstrained handwritten numerals with wavelet transform and multilayer cluster neural network,'' \emph{Pattern Recognit.}, vol.~29, no.~12, pp. 1953--1961, 1996.

\end{thebibliography}

\end{document}